\documentclass[lettersize,journal]{IEEEtran}
\usepackage{amsmath,amsfonts}
\usepackage{algorithm}
\usepackage{algpseudocode}
\usepackage{array}
\usepackage[caption=false,font=normalsize,labelfont=sf,textfont=sf]{subfig}
\usepackage{textcomp}
\usepackage{stfloats}
\usepackage{url}
\usepackage{verbatim}
\usepackage{graphicx}
\usepackage{cite}
\usepackage[table]{xcolor}
\usepackage{multirow} 
\def\BibTeX{{\rm B\kern-.05em{\sc i\kern-.025em b}\kern-.08em
    T\kern-.1667em\lower.7ex\hbox{E}\kern-.125emX}}
\usepackage{balance}
\begin{document}
\title{Adapting Mental Health Prediction Tasks for Cross-lingual Learning via Meta-Training and In-context Learning with Large Language Model}
\author{Zita~Lifelo,
        Huansheng~Ning,~\IEEEmembership{Senior Member, IEEE}
        and Sahraoui~Dhelim
\IEEEcompsocitemizethanks{\IEEEcompsocthanksitem Z. Lifelo is with the School
of Computer and Communication Engineering, University of Science and Technology Beijing, Beijing, China.\protect\\
E-mail: zitalife@gmail.com
\IEEEcompsocthanksitem S. Dhelim is with the School of Computer Science, University College Dublin, Dublin 4, Ireland.\protect\\
E-mail: sahraoui.dhelim@ucd.ie
\IEEEcompsocthanksitem H. Ning is with the School of Computer and Communication Engineering, University of Science and Technology Beijing, Beijing, China.\protect\\
E-mail: ninghuansheng@ustb.edu.cn}
\thanks{Manuscript received April 19, 2024; revised August 26, 2024.}}

\markboth{Journal of \LaTeX\ Class Files,~Vol.~XX, No.~X, August~2024}%
{Shell \MakeLowercase{\textit{et al.}}: Your Research Paper Title}


\maketitle


\markboth{MANUSCRIPT ID NUMBER}%
{How to Use the IEEEtran \LaTeX \ Templates}

\maketitle

\begin{abstract}
Timely identification is essential for the efficient handling of mental health illnesses such as depression.  However, the current research fails to adequately address the prediction of mental health conditions from social media data in low-resource African languages like Swahili. This study introduces two distinct approaches utilising model-agnostic meta-learning and leveraging large language models (LLMs) to address this gap. Experiments are conducted on three datasets translated to low-resource language and applied to four mental health tasks, which include stress, depression, depression severity and suicidal ideation prediction. we first apply a meta-learning model with self-supervision, which results in improved model initialisation for rapid adaptation and cross-lingual transfer. The results show that our meta-trained model performs significantly better than standard fine-tuning methods, outperforming the baseline fine-tuning in macro F1 score with 18\% and 0.8\% over XLM-R and mBERT. In parallel, we use LLMs' in-context learning capabilities to assess their performance accuracy across the Swahili mental health prediction tasks by analysing different cross-lingual prompting approaches. Our analysis showed that Swahili prompts performed better than cross-lingual prompts but less than English prompts. Our findings show that in-context learning can be achieved through cross-lingual transfer through carefully crafted prompt templates with examples and instructions.

\end{abstract}


\section{Introduction}
\IEEEPARstart{D}{epression}  has a profound impact on mental health, which is already significantly influenced by stress and anxiety. According to the World Health Organisation (WHO) \cite{WHO2020Stats}, this disorder, characterised by enduring feelings of melancholy and a diminished capacity for interest or concentration, impacts more than 264 million individuals worldwide and is a significant contributor to disability and suicide. Before the exacerbation of these problems by the COVID-19 pandemic, over 116 million people in Africa were already experiencing mental health disorders \cite{WorldMentalHealth2022}, with almost 29 million persons affected by depression \cite{Gbadamosi2022}. In the current age of living in cyberspace, advancements in social media and online platforms have facilitated the early identification of mental disorders such as depression \cite{mentaldistress} through the analysis of social media data \cite{Choudhury2017,suicide,hatespeech}. This has proven beneficial for researchers in comprehending patterns related to mental health.

Although Swahili is widely spoken in Africa, there is a lack of research on predicting mental health disorders in low-resource African languages. The growing amount of Swahili digital content on the internet requires sophisticated automatic language technologies. Despite using Swahili online data for tasks such as sentiment analysis \cite{Martin2021}, the complexity of mental health illnesses like depression or suicide and the laborious data collection and annotation process impede the development of mental health datasets, challenging the training of effective deep learning models for Swahili.

Recent advancements in large language models (LLMs) like OpenAI's GPT and deep learning techniques, including CNN, Bi-LSTM, and transformer-based models like mBERT and XLM-RoBERTa (XLM-R), have significantly advanced natural language processing (NLP) tasks in mental health as seen in studies \cite{Yang2023,Lamichhane2023,Ghanadian2023,Xu2023,Amin2023}.  Furthermore, researchers have also created multilingual generative language models that perform similarly to GPT-3 for multilingual tasks involving few-shot learning and machine translation \cite{Lin2022}. These models show some potential benefits for low-resource languages. Still, their success depends on significant computational resources and large datasets, usually lacking in low-resource languages like Swahili. The challenge lies in adapting these advanced models for effective mental health prediction with limited or no specific data available.

Furthermore, researchers are adopting multilingual approaches to tackle the limited availability of annotated data in low-resource languages \cite{Vajrobol2023,B2019}. These methods frequently involve utilising resources from a high-resource language, such as English, to improve a low-resource language through cross-lingual transfer. Machine Translation (MT) techniques offer another pathway by translating texts between languages \cite{Cattoni2021,Popel2020}, aiding in multilingual tasks such as depression prediction. Pre-trained language models, fine-tuned to high-resource languages, can provide shared embeddings for multiple languages, facilitating cross-lingual learning. Zero-shot and few-shot learning strategies primarily focus on addressing the challenges of low-resource languages. However, fine-tuning these models proves difficult due to the scarcity of data in these languages.

In addition, meta-learning, which provides a solid basis for fine-tuning models \cite{Awal2023}, utilising zero-shot and few-shot learning, is beneficial when training scarce data. This strategy is especially advantageous for improving efficiency in languages with limited resources. In addition, LLMs, which have undergone extensive training on large datasets, exhibit near human-writing capabilities and have demonstrated outstanding performance in various benchmarks, including mental health \cite{Liu2023}. The models' capacity to understand complex concepts and perform tasks such as mental health classification, even with minimal direct examples, showcases their exceptional abilities. Additionally, instruction tuning (IT) enables these models to quickly adapt to specific domains without requiring extensive training or modifications \cite{Wei2022}, improving their adaptability.

Our contributions are summarized as follows:
\begin{itemize}
    \item We propose two frameworks for cross-lingual transfer and adaptation of mental health prediction tasks in a low-resource language: a few-shot cross-lingual model agonistic meta-learning framework and a cross-lingual transfer and adaptation using LLM via in-context learning. 
    \item We conduct a cross-lingual meta-training using the English-Swahili language pair. We investigate whether meta-training is better than standard multilingual fine-tuning when training instances are only available for some languages and when both high and low-resource languages have sufficient training data. Furthermore, by examining cross-lingual transfer through templates and instructional examples, we conduct an in-depth analysis of three cross-lingual prompting approaches: Swahili, Cross-lingual, and English.
    \item We apply these two methods to four mental health prediction tasks: stress prediction, depression prediction, depression severity prediction and suicidal ideation prediction using datasets translated into Swahili. We evaluate LLM performance across these prompt approaches and assess the impact of non-English prompts on mental health prediction tasks.
\end{itemize}

We organize the rest of the paper as follows: Section II provides an overview of the related work. Section III presents our study's methodology. Section IV discusses the datasets used and the mental health prediction tasks. Section V describes our experiments, and the results are discussed and analysed in Section VI. We highlight the conclusion in Section VII. 

\section{Related Work}
The use of NLP techniques has advanced the prediction of mental health on social media platforms. This progress has been made possible by developing deep learning and transformer-based models. Transformer-based models have become crucial in NLP tasks, especially for high-resource languages like English. This is due to their effective attention-based transformer design, as introduced by Vaswani et al. \cite{Vaswani2017}, and the masking pre-training procedures introduced by BERT \cite{Devlin2019}. The application of these approaches has greatly improved the performance of models in multiple studies \cite{Hoffmann2022,Chowdhery2022,Thoppilan2022,Brown2020,Rae2021}demonstrating substantial progress in the field.

Although there have been significant developments in language technology, there is a noticeable lack of attention given to low-resource languages \cite{Joshi2020}, which are underrepresented among the world's roughly 7,000 languages. Language technologies play a vital role in upholding linguistic diversity worldwide and encouraging the use of multiple languages. Thus, it is crucial to prioritise research on cross-lingual and multilingual NLP, meta-learning, and developing language models to improve mental health prediction in underrepresented languages on online platforms.

Cross-lingual learning is crucial in NLP as it allows knowledge transfer from languages with high resources to low resources languages. This is particularly beneficial for extremely low-resource languages. Studies, such as the one conducted by Urbizu et al. \cite{Urbizu2023}, have demonstrated that language models pre-trained using machine-translated data can perform well. This suggests that this approach could be an effective technique for improving NLP applications in low-resource languages. Ghafoor et al.'s \cite{Ghafoor2021} study on dataset translation demonstrates the efficacy of multilingual text processing in addressing the challenges encountered by low-resource languages, indicating its wide-ranging application and success in various linguistic contexts. Although there have been just a few studies on predicting mental health using cross-lingual and multilingual approaches, the current research utilises model fine-tuning and translation methods to address the lack of data \cite{Vajrobol2023,B2019}, \cite{Munthuli2023,Aulia2022,Pool-Cen2023}.

Meta-learning, often known as learning to learn, emphasises the development of models that can rapidly adapt to new tasks with minimal examples. This approach diverges from conventional supervised learning by building a flexible foundation for quick adaptation to new tasks. Meta-learning encompasses two significant techniques: optimization-based \cite{Finn2017} and metric-based techniques \cite{Gordon2018,Koch}. For instance, Finn et al. \cite{Finn2017} created the Model-Agnostic Meta-Learning (MAML) framework, which was specifically developed to enhance efficiency in instances when there is a need for learning from only a few examples. In addition, the usefulness of meta-learning can be applied to cross-lingual applications, as seen in the work of Nooralahzadeh et al. in their X-MAML \cite{Nooralahzadeh2020}, which focuses on improving natural-language understanding in different languages. Furthermore, meta-learning has demonstrated potential in identifying hate speech and offensive speech in different languages \cite{Awal2023, Mozafari2022} highlighting its extensive usefulness in addressing a wide range of complex NLP tasks. However, no study has been conducted using meta-learning for multilingual or cross-lingual mental health prediction tasks. Dhelim et al. focused on using the users' persoanlity traits to optimize the accuracy of recommendation systems \cite{personet,productRec,personalitySurvey,personalityhybrid} and user interest mining \cite{compath,KBS_personality}.

The development of large language models like the GPT has significantly advanced NLP applications. The GPT-3 model \cite{Brown2020}, by Brown et al., has been recognized for its exceptional few-shot learning capabilities. In addition, InstructGPT \cite{Ouyang} employs a reinforcement learning from human feedback (RLHF) technique, enhancing performance and instruction adherence. Subsequent iterations, such as GPT-3.5-turbo (ChatGPT) and GPT-4, incorporate chat-based interactions in their training, further refining their responsiveness and contextual understanding \cite{Kalyan2023}.  These models have demonstrated strong zero-shot capabilities in various NLP tasks \cite{Laskar2023,Bang2023,Qin2023,Yang2023}, yet their application in multilingual mental health analysis remains underexplored. Research on GPT models, including their multilingual functionalities, indicates they outperform traditional methods but face challenges with African languages \cite{Rathje}. Although GPT-4 demonstrates high ability in multiple languages, it still has limits in understanding other complex languages and translating accurately. This indicates the need for future improvements to maximise its usefulness in multilingual tasks.

\section{Methodology}
In incorporating an established methodology into our research, we acknowledge the importance of utilising known methodologies while simultaneously addressing the unique challenges posed by our specific area of study. The methodology employed in this study, based on MAML and LLMs, has been previously used in various NLP domains. However, its application in cross-lingual transfer and adaptation for mental health prediction tasks, particularly in low-resource language (Swahili), presents a novel context.

\subsection{Meta-Learning}
Our approach is designed to improve the cross-lingual transfer in low-resource target languages. We generate a set of tasks by utilising instances from both the source and target languages. Next, we simulate the recurrent meta-training procedure similar to MAML.  The objective is to enhance the initialization of model parameters, enabling adaptability to low-resource target language challenges by using a limited number of labelled data, referred to as few-shot learning.  We employ meta-learning methodologies to train the model on the target language. 

\subsubsection{Model Agnostic Meta-Learning}
The MAML algorithm is designed to assign a probability distribution \(p(\mathcal{T})\) to a set of tasks \(\{\mathcal{T}_1,\mathcal{T}_2,\ldots,\mathcal{T}_M\}\). The meta-model \(f_\theta(\cdot)\) with parameters \(\theta\) is learned through iterative meta-training on a set of sample tasks \(\mathcal{T}\) that are selected from a probability distribution \(p(\mathcal{T})\), i.e., \(\mathcal{T}_i \sim p(\mathcal{T})\). The MAML framework comprises an inner loop that enables task-specific adaptation, as well as an outside loop that facilitates quick adaptation to new and unfamiliar tasks through the process of meta-learning.

MAML optimises the meta model \(f_\theta\) for a specific task \(\mathcal{T}_i\) via gradient descent. The updated model parameters \(\theta_i'\) are obtained by subtracting the product of the step size \(\alpha\) and the gradient of the classification loss \(\mathcal{L}_{\mathcal{T}_i}\). The evaluation of the task-specific training outcome is then performed on the corresponding test set.

\begin{equation}
\theta_i' \gets \theta - \alpha \nabla \mathcal{L}_{\mathcal{T}_i}(f_\theta)
\end{equation}

The primary goal of the meta-learner optimisation is to minimise the meta loss, which is calculated by considering all the training tasks.

\begin{equation}
\underset{\theta}{\text{min}} \sum_{i=1}^{m} \mathcal{L}_{\mathcal{T}_i}(f_{\theta'}) = \sum_{i=1}^{m} \mathcal{L}_{\mathcal{T}_i}(f_{\theta-\alpha \nabla_\theta \mathcal{L}_{\mathcal{T}_i}(f_\theta)})
\end{equation}

The meta parameter is then updated to:

\begin{equation}
\theta = \theta - \beta \nabla_\theta \sum_{i=1}^{m} \mathcal{L}_{\mathcal{T}_i}(f_{\theta'})
\end{equation}

This is achieved by minimising the sum of the individual losses of each task. Here, \(\beta\) represents the learning rate of the meta-learner. 

\begin{figure}[htbp]
  \centering
  \includegraphics[width=\linewidth]{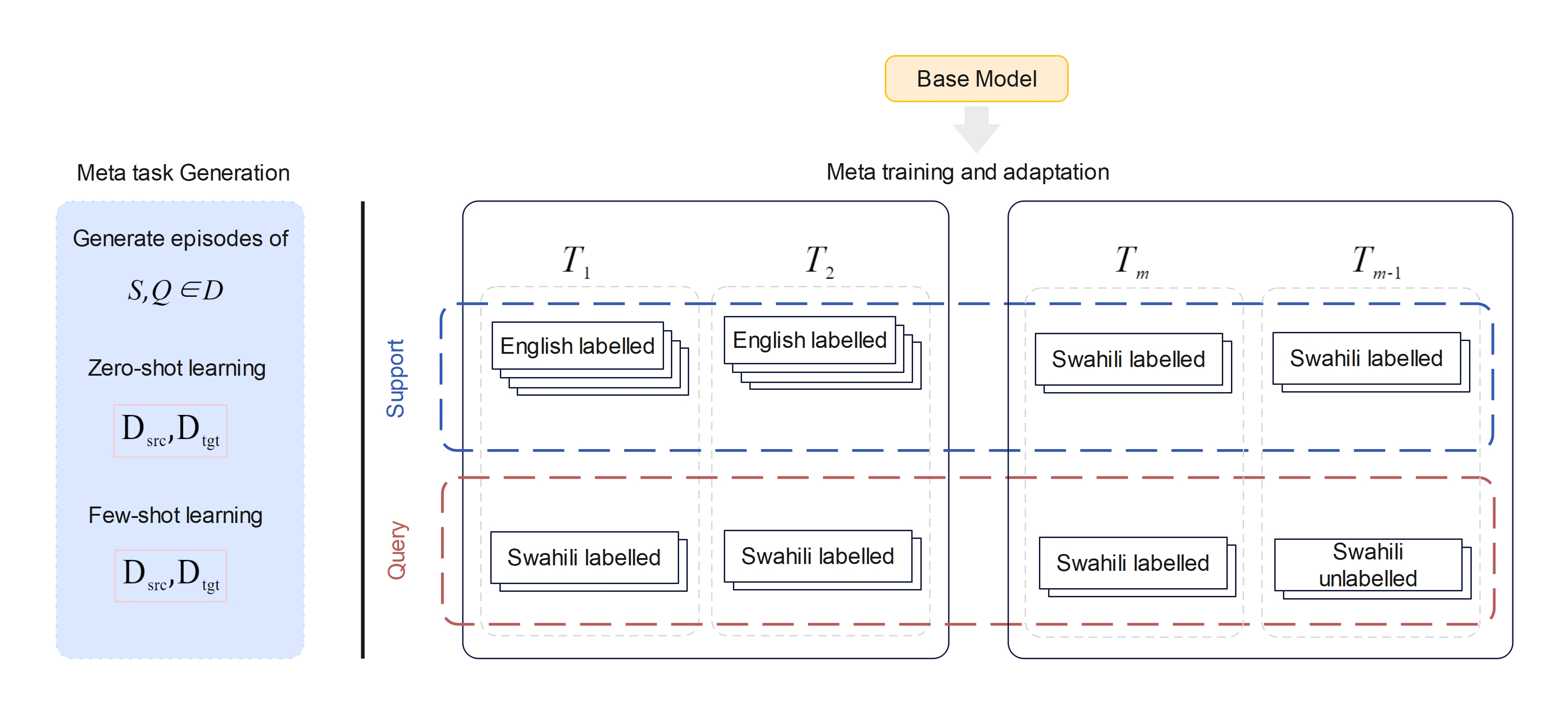}
  \caption{An overview of the proposed framework: we use English as the source and Swahili as the target language. The meta-train stage transfers from the source to the target languages, while the meta-adaptation further adapts the model to the target language.}
  \label{fig_framework}
\end{figure}

To update \(\theta\), we can accumulate multiple instances of tasks. In reality, a first-order approximation is used for MAML algorithm training, even though it requires double gradient descent optimisation. Meta-learning serves as a good foundation for promptly carrying out the process of learning a new task.
\subsubsection{Proposed MAML Model}
We use optimization-based meta-learning, akin to the approach used in~\cite{Awal2023},~\cite{Mhamdi2021}. We employ two pre-trained models, mBERT and XLM-R, with two levels of adaptation: (i) meta-training from the source language to the target language (ii) meta-adaptation on the same target language for more language-specific adaptation as shown in Fig~\ref{fig_framework}. To improve cross-lingual transfer, we use a few labelled samples (few-shot learning) in the target language which helps with quick adaptation. We apply our approach to the mental health prediction tasks described in Table ~\ref{tab_datasets}, of cross-lingual prediction in target language.

Applying meta-learning to a task requires the construction of multiple pseudo-tasks, which are instantiated as pairs of datasets. In the sections that follow, we describe these training datasets and meta-task generation. Finally, we present our algorithm, Cross-lingual Mental Prediction MAML, in detail.

\begin{enumerate}
  \item \textbf{MAML model Training Data:} A set of support and query batches sampled from $\mathcal{D}$ where $\mathcal{D}$ refers to available training data from training languages is required. Suppose we have samples in source $\mathcal{D}_{\text{src}}$ and target $\mathcal{D}_{\text{tgt}}$ languages. Our training data $\mathcal{D}$ consists of the tuple: $(\mathcal{D}_{\text{src}}, \mathcal{D}_{\text{tgt}})$ in both the zero-shot and the few-shot settings. We then split the training set for each language into support ($\mathcal{D}_{\text{lang}}^{\text{support}}$) and query ($\mathcal{D}_{\text{lang}}^{\text{query}}$), where $\text{lang} \in \{\text{src}, \text{tgt}\}$.
  \item \textbf{Meta-Tasks Generation:} The model requires episodic training on meta-tasks containing support (meta-training) and query (meta-testing) sets. For each episode, two subsets of examples are sampled separately: the support set $\mathcal{S}$ and the query set $\mathcal{Q}$. We repeat this process multiple times.
\end{enumerate}
\subsubsection{Cross-lingual Mental Prediction MAML Model Algorithm}
Following the notation described in the sections above, we present our algorithm. Model training requires a base classifier. Any multilingual encoder such as mBERT, XLM-R fine-tuned on a source language can be used. The training choice is passed to the Algorithm along with the base model parameters. Then, the meta-learner parameters $\theta$ are initialized from a fine-tuned base model. Our adaptation of MAML to cross-lingual transfer learning is in two stages. We sample a batch of tasks $\mathcal{T}_i$ from distribution $\mathcal{D}$. For every task in $(\mathcal{S}, \mathcal{Q})$, we take one inner loop gradient step using training loss on $\mathcal{S}$ and adapt model parameters to $\theta'$. Then, we evaluate task-specific training outcomes using $\mathcal{Q}$, and save it for the outer loop update. Each task has its own task-adaptive model parameters $\theta_i'$. At the end of all tasks of each batch, we sum over all pre-computed gradients and update $\theta$, thus completing one outer loop. The difference between meta-train and meta-adapt stages comes down to the parameters and hyperparameters passed into Algorithm 1.
\begin{algorithm}
\caption{Cross-lingual Mental MAML}
\begin{algorithmic}[1]
\setcounter{ALG@line}{0}
\Statex \textbf{Input:} Training choice $\mathcal{C}$, high-resource language $\mathcal{H}$, low-resource language $\mathcal{L}$, Task set distribution $\mathcal{D}$, Meta-learner $\mathcal{M}$ with parameters $(\theta,\alpha,\beta)$
\State Fine-tune $\theta$ on $\mathcal{H}$ (base model)
\State Initialize $\theta \gets \theta$
\If{$\mathcal{C}$ is zero-shot}
    \State Select target language from $\mathcal{L}$ to get $\mathcal{D} = \mathcal{D}_{\text{src}}, \mathcal{D}_{\text{tgt}}$
\Else
    \State Select target language get $\mathcal{L}$ to form $\mathcal{D} = \mathcal{D}_{\text{src}}, \mathcal{D}_{\text{tgt}}$
\EndIf
\While{not done}
    \State Sample batch of tasks $\mathcal{T} = \{\mathcal{T}_1, \mathcal{T}_2, \ldots, \mathcal{T}_M\} \sim \mathcal{D}$
    \ForAll{$\mathcal{T}_i = \{\mathcal{S}, \mathcal{Q}\}$ in $\mathcal{T}$}
        \State Evaluate $\nabla_\theta \mathcal{L}_{\mathcal{T}_i}(\mathcal{M}_\theta)$ on $\mathcal{S}$
        \State Evaluate adapted parameters with gradient descent: $\theta' = \theta - \alpha \nabla_\theta \mathcal{L}_{\mathcal{T}_i}(\mathcal{M}_\theta)$
        \State Evaluate $\mathcal{L}_{\mathcal{T}_i}(\mathcal{M}_{\theta'})$ using $\mathcal{Q}$ for outer loop meta-update
        \State Update $\theta \gets \theta - \beta \nabla_\theta \sum_{i} \mathcal{L}_{\mathcal{T}_i}(\mathcal{M}_{\theta'})$
    \EndFor
\EndWhile
\State Evaluate the test set on the target language using $\mathcal{M}_\theta$
\end{algorithmic}
\end{algorithm}

In meta-training, the task sets are sampled from the distribution which uses high-resource (typically English) data in support sets and low-resource data in the query sets. In the meta-adapt stage, the task sets are sampled from the distribution which uses low-resource data in both support and query sets. This ensures the model knows how to learn from examples within the target language under a low-resource scenario. 
\subsection {Cross-lingual transfer and adaptation using Large Language Model}

In this section, we introduce our LLM cross-lingual transfer and adaptation for the in-context learning framework. We provide a description of the methodology setup, then, we outline three different prompt designs for generating input-output pairs for task-specific information from our datasets. 
\subsubsection{In-context learning}

We adapt the learning framework as proposed by Brown et al.~\cite{Brown2020} and similar to~\cite{Lin2022} for our methodological settings. We let $\mathcal{M}$ be a large language model, fine-tuned to effectively handle tasks $\mathcal{D}$. $\mathcal{D} = (\mathcal{P}, \mathcal{E})$ comprises a task description $\mathcal{P}$ and a few demonstration examples in one or more languages
\[
\mathcal{E} = \bigcup_{l=1}^{|\mathcal{L}|} \mathcal{E}^l
\]
The task description $\mathcal{P}$ combines a template $\mathcal{T}$ with a verbalizer $v$ as a prompt $\mathcal{P} = (\mathcal{T},v)$. $\mathcal{T}$ maps an input example $x$ to a string $\mathcal{T}(x)$ that includes a [Mask]. For classification tasks, $v\colon \mathcal{Y} \rightarrow \mathcal{V}^*$ is a verbalizer that converts each candidate label or choice $y \in \mathcal{Y}$ into a string $v(y)$. Both $\mathcal{T}(x)$ and $v(y)$ can be divided into a series of one or more tokens from the language model vocabulary $\mathcal{V}$. An instantiated prompt $\mathcal{P}(x, y)$ is created by replacing the [Mask] symbol in $\mathcal{T}(x)$ with $v(y)$.

Zero-shot learning: For a given test example $\widetilde{x}^t$ in a target language $t$, the zero-shot prediction $y$ is the output that maximizes the language model-based scoring function $\sigma$, which is calculated over the mental health prediction tasks.
\begin{equation}
\hat{y} = \underset{y}{\mathrm{argmax}}\ \sigma\left(\mathcal{M}, \mathcal{P}(\widetilde{x}^t,y)\right)
\end{equation}
In this instance, the variable $y$ iterates over the various classifications of tasks.

Few-shot prompting: Suppose you have a source language with $k$ demonstration examples, represented as
\[
\mathcal{E}^s = \left\{(x_i^s,y_i)\right\}_{i=1}^k
\]
In this instance, $x_i^s$ refers input (e.g., social media post) and $y$ to its classification or label. Concatenating the created prompts of the demonstration examples $\left\{\mathcal{P}(x_i^s,y_i)\right\}_{i=1}^k$ and using it as a context prefix for the input string used in zero-shot setting allows few-shot learning:
\begin{equation}
\hat{y} = \underset{y}{\mathrm{argmax}}\ \sigma\left(\mathcal{P}(x_1^s,y_1)[\text{sep}]\ldots\mathcal{P}(x_k^s,y_k)[\text{sep}]\mathcal{P}(\widetilde{x}^t,y)\right),
\end{equation}
where $[\text{sep}]$ is separator symbol chosen empirically. When $s = t$, we have the in-language few-shot learning setup. When $s \neq t$, we have the cross-lingual few-shot learning setup.

\begin{figure*}
\centerline{
\includegraphics[width=7in]{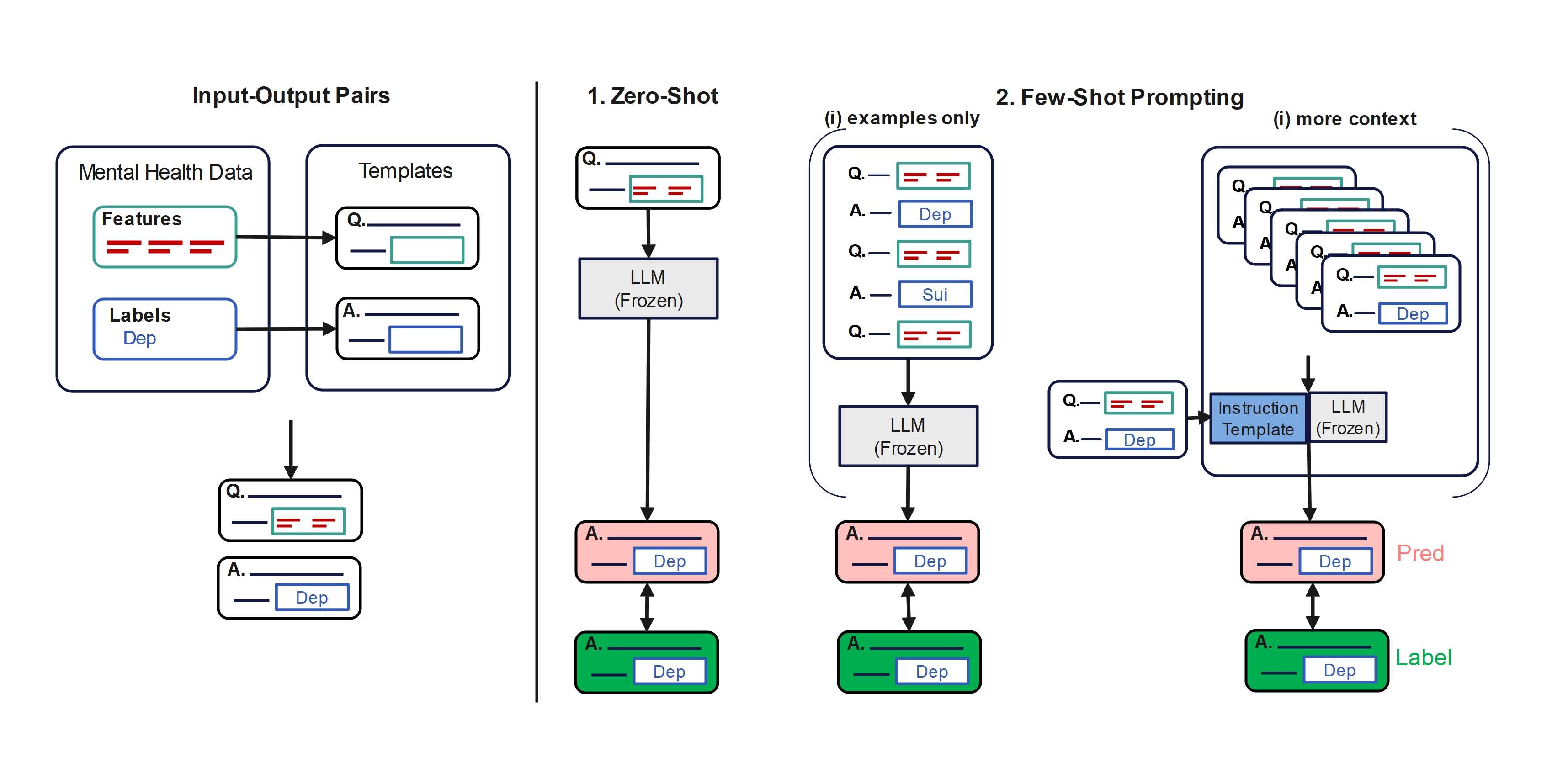} 
}
\caption{LLM in-context settings. We use psychological data and social media datasets to construct input-output pairs. We create training and test splits, comparing zero-shot and few-shot settings models' prediction performance.}
\label{fig_methodology}
\end{figure*}
The summary of settings for our methodology is presented in Fig~\ref{fig_methodology}. For our method, we divided few-shot setup into two scenarios: (i) where we only provide examples of the input-output pair in the prompt and (ii)where we provide more context by including additional instructions or demonstrations on how the tasks should be carried out within the prompt. For each mental health classification task, we embed the textual features from our datasets into textual templates and combine with psychological data to construct input-output pairs.

\subsubsection{Formulation of Input-Output Pairs}

We create a set of tasks tailored to the detailed needs of our methodology. These tasks include \textit{stress prediction}, \textit{depression prediction}, \textit{depression severity prediction}, and \textit{suicide ideation prediction}, shown in Table ~\ref{tab_datasets}. Formulating the input-output pairs follows a structured sequence-to-sequence format that includes the definition of the tasks describing each task's goal and intended outcome. Instance inputs which are instances obtained from datasets that represent scenarios concerning mental health. Constraints that provides specific guidelines that direct how the task should be carried out. A prompt which acts as an instruction that links the task definition to its execution. Then the expected output which is designed to be logical and relevant.
\begin{figure*}
\centerline{
\includegraphics[width=7in]{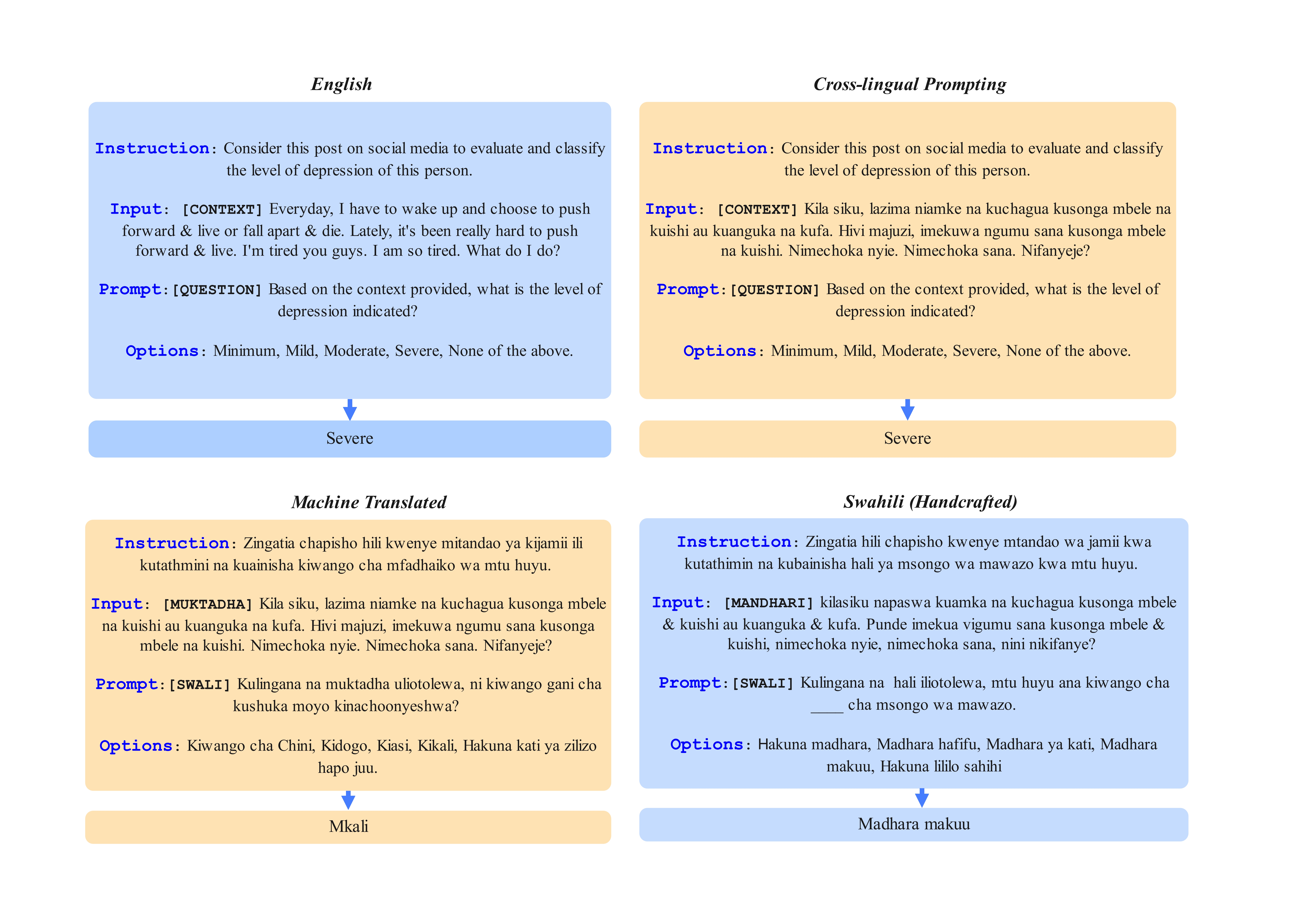} 
}
\caption{An example of an instruction based input-output pairs for the depression severity prediction task using four prompting strategies. The task is formatted as a natural language sequence. Each input contains an instruction, instance and task specific prompts.}
\label{fig_prompt-design}
\end{figure*}

We adopt a cross-lingual prompt design framework using the predetermined tasks by considering three techniques for creating prompts:
\begin{enumerate}
    \item Swahili Prompts: engaging native speakers to ensure natural language use. Although ideal, this method is resource-intensive.
    \item Machine Translated Prompts: employing automated systems to translate prompts into our target language and apply the prompts directly.
    \item Cross-lingual Prompting: Using the language model's cross-lingual abilities acquired through training on a variety of language data, this technique applies English prompts to non-English examples.
\end{enumerate}
Fig~\ref{fig_prompt-design} shows an example of an instance for the depression severity prediction task in the three different prompt settings. To improve the model's ability to generalise, we construct the prompt in a manner that enables the model to choose the suitable instruction that corresponds to the given input-output pair and evaluate if a given instruction results in the correct output. Each task's formatting also involves using special tokens such as [CONTEXT] and [QUESTION] to define distinct sections of the input and intended output properly.

\section{Datasets}
Our experiment is based on three well-known datasets used for mental health analysis, obtained from the Reddit. We chose these three datasets since they were supervised and annotated by human experts. We define four diverse mental health prediction tasks based on these datasets. Table ~\ref{tab_datasets} provides a summary of the datasets and tasks. The datasets are then translated into the Swahili language using Google Cloud translation API\footnote{https://cloud.google.com/translate}\footnote{The datasets were translated as we could not find any mental health datasets specifically for the Swahili language. Our primary source of datasets is the Reddit platform, which provides high-quality labels and is chosen for its accessibility. Obtaining other good-quality datasets is challenging because they are confidential. We will address this within the limitations.}. The translation of the datasets was further scrutinized and evaluated by three Swahili native speakers to confirm linguistic and contextual accuracy.

\begin{table*}[!htbp]
\captionsetup{size=footnotesize} 
\caption{Summary of Datasets and mental health prediction tasks}
\label{tab_datasets}
\centering
\setlength{\tabcolsep}{12pt} 
\renewcommand{\arraystretch}{1.2} 
\footnotesize 
\begin{tabular}{|c|c|p{4cm}|c|} 
\hline
\rowcolor{lightgray} \textbf{Dataset} & \textbf{Task} & \textbf{Example prompt in English} & \textbf{Target} \\ \hline
Dreaddit & Binary Stress Prediction (Task 1) & Determine if the poster of this post is [Stress]. & [Stress]  \\ \hline
\multirow{2}{*}{DepSeverity} & Binary Depression Prediction (Task 2) & Based on the context provided, determine which level of [Depression] risk the poster has. & [Depression] \\ \cline{2-4} 
 & Depression Severity Prediction (Task 3) & Based on the context provided, determine which level of [Depression] risk the poster has. & [Depression] \\ \hline
SDCNL & Binary Suicide Ideation Prediction (Task 4) & Determine if the poster of this post want to [Suicide]. & [Suicide] \\
\hline
\end{tabular}
\end{table*}

\begin{enumerate}
    \item \textbf{Dreaddit \cite{Turcan2019}:} The dataset was obtained from Reddit and included posts from ten subreddits spanning five domains: abuse, social, anxiety, PTSD, and wealth.  Several human annotators analysed whether language portions reflected the poster's stress, and the comments were integrated to create final labels. We use this dataset for binary stress detection as task 1.
    \item \textbf{DevSeverity \cite{Naseem2022}:} This dataset utilised the identical postings collected in \cite{Turcan2019}, although with a specific focus on depression. Two human annotators adhered to the DSM-5 guidelines and categorised the postings into four distinct levels of depression: minimum, mild, moderate, and severe. This dataset is utilised for task 2, which involves binary depression prediction, specifically determining whether a post contains at least mild depression. Additionally, it is used for task 3, which focuses on predicting the degree of depression.
    \item \textbf{SDCNL \cite{Haque2021}:} This dataset also gathered postings from Reddit, from the subreddits r SuicideWatch and r/Depression, contributed by 1723 members. Using manual annotation, the individuals categorised each post based on whether it exhibited indications of suicide ideation. This dataset is utilised for the task of binary suicide ideation prediction, as task 4.
\end{enumerate}
For differentiating between the main English datasets and the translated Swahili datasets, we will use a subscript ‘sw’ against each name of the dataset in our tables.  

\section {Experiment}
\subsection{Settings}
In the meta-learning experiment, we employ three training configurations: zero-shot without fine-tuning in the target language, domain adaptation where we train on English and Arabic datasets to evaluate cross-lingual transfer, and full fine-tuning by utilising training samples from three languages. We employ English samples to train the base model. The meta-training samples are obtained from the validation sets of the source languages and the training sets of the target languages. The training sets of the source languages are utilised to train the base model. In all the experiments, we combine samples from each language and then create meta-tasks for meta-training. Using 32 shots, we sample task triplets.
For LLM, we use the gpt-3.5-turbo and gpt-4 model. The models are closed-source and available through API provided by OpenAI. We evaluate the models on zero-shot and few-shot settings. We also evaluate three approaches to prompting as discussed in section III. We randomly draw examples from the training data and report the average performance across 5 runs, randomly sampling a different set of few-shot examples each time. We report evaluation results on the test sets and pre-select a few k values with 0 and 4 shots. We benchmark zero-shot learning performance and report 4-shot performance.
To implement the transformer-based models, we use the HuggingFace transformers library. The selected pre-trained model is initialised using pre-trained weights given in the transformers library.  The classification heads are implemented using the PyTorch initialised with random values. We use a NVIDIA GeForce RTX 3060 GPU. For evaluation metrics, we use accuracy for the LLM experiment and macro F1 score for the meta-learning experiments.
\subsection {Baselines}
Standard LM fine-tuning is adapted as a baseline for all for our meta-learning experiments. We fine-tune on mBERT and XLM-R. We evaluate the LLM models in-context learning settings and benchmark our zero-shot experiments.

\section {Results and Analysis}
We summarise our experimental results for cross-lingual transfer and adaptation on mental health tasks using meta-learning and LLM approaches. 
\subsection {Meta-training Evaluation}
We focus on the macro F1-score, the average result from five random iterations for all our meta-learning experiments.
\subsubsection{Zero-shot Meta-training Experiment}
Table~\ref{tab:meta-learning-results} presents the outcomes of zero-shot experiments, with mBERT and XLM-R initializations denoted by the symbols † and ‡, respectively. Baseline models exhibit poor performance when it comes to zero-shot scenarios. Our meta-learning zero-shot models surpass baseline models with over 20 percent in performance, demonstrating satisfactory results compared to mBERT and XLM-R models. Models initialised with XLM-R surpass those initialised with mBERT. This may be attributed to the substantial pre-training of XLM-R on a vast corpus and its complex architecture. XLM-R constantly exhibits a superiority across all tasks.

For datasets such as ours, which heavily rely on translated text, self-training is a valuable method for enhancing the performance of trained models. It decreases the need for a labelled training set in the target language \cite{Awal2023}. Self-training our proposed model yields comparable performance to meta-training in a zero-shot scenario. The increase in performance is achieved by using meta-training on silver labels from the target, and iteratively improving the model through self-refinement. To achieve a balance between noise and ground truth, we employ a validation set in the source language during the self-training phase of our model. We observe an enhancement in performance when using accurate ground truth labels throughout training.
\begin{table}[htbp]
  \centering
  \caption{Meta-learning zero-shot evaluation of target low resource language on four mental health tasks.}
  \label{tab:meta-learning-results}
  \includegraphics[width=\linewidth]{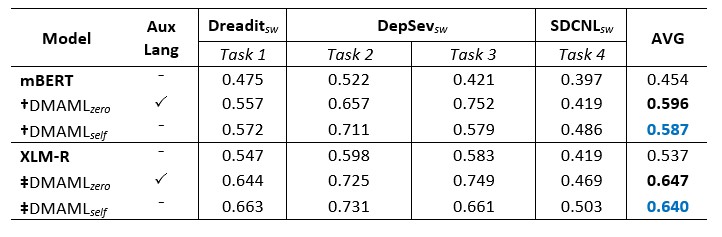}
\end{table}

\begin{table}[htbp]
  \caption{Domain adaptation experiments on both fine-tuning and meta-training. We use languages English and Arabic for training and evaluate on all the Swahili language mental health tasks.}
  \label{tab:domain-adaptation-experiments}
  \centering
  \includegraphics[width=\linewidth]{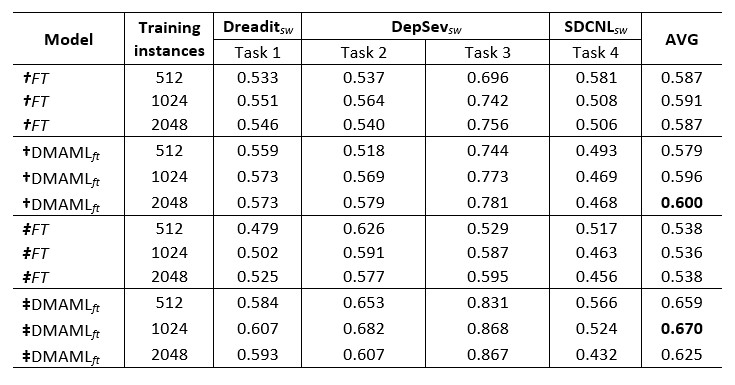}
\end{table}
\subsubsection{Domain Adaptation}
Table~\ref{tab:domain-adaptation-experiments} summarises our domain adaptation experiments, where we used English and Arabic \cite{Nassar2022} datasets as auxiliary training datasets. The trained model is then evaluated on the Swahili test dataset. The number of training instances was adjusted to account for dataset restrictions and varied between 512, 1024 and 2048. We used the maximum available for training for datasets with fewer instances. Our adapted DMAMLft model showed comparable performance compared to the fine-tuning, especially for XLM-R outperforming the fine-tuned baseline with an overall of 25 percent improvement while training 1024 instances. The domain-adaptive training helps to retain consistent performance on source and target language compared to standard fine-tuning.

\subsubsection{Fine-tuning vs Meta-training}
\begin{figure*}
\centerline{
\includegraphics[width=\textwidth]{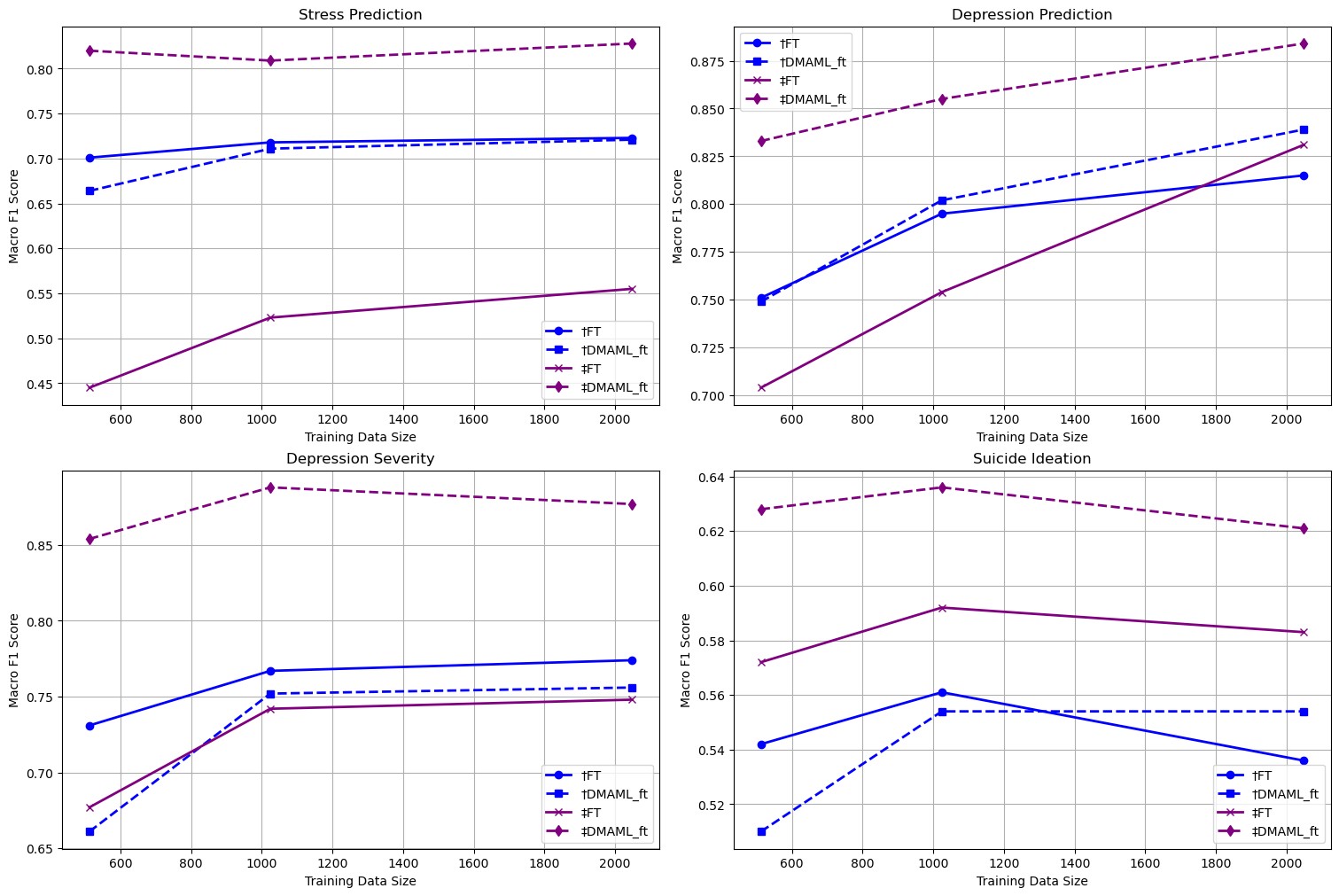}  
}
\caption{Fine-tuning and meta-training on mental health tasks. We used training sets to train on English, Arabic and Swahili then fine-tuned aggregated data, varying training instances. Blue lines indicate mBERT initialised and purple XLM-R initialised.}
\label{fig_mental-health-tasks}
\end{figure*}

We also evaluated our meta-training strategy while training on data in English, Swahili and Arabic. We varied the training data instances and evaluated the performance under increasing data availability. We create a set of meta-tasks by gathering support and query sets from the three languages and then fine-tune the aggregated training data. Our findings depicted in Fig ~\ref{fig_mental-health-tasks} show that our meta-trained model performs significantly better than standard fine-tuning methods, outperforming the baseline fine-tuning with 0.8 percent and 18 percent over mBERT and XLM-R using 2048 instances, respectively. These results highlight the importance of meta-training in improving cross-lingual transfer in mental health prediction tasks.


\subsection {Large Language Model Evaluation}
\subsubsection{In-context Learning}
We evaluate large language models (gpt-3.5-turbo and gpt-4) in-context learning for our mental health tasks. We randomly draw examples from the training data and report the average performance accuracy across 5 runs for few-shot and in-context learning, randomly sampling a different set of few-shot examples each time. We report evaluation results on the test sets and pre-select a few k values with 0 and 4 shots. We benchmark zero-shot learning performance and report 4-shot performance. We use few-shot examples in the Swahili language unless otherwise specified.
We examine the capabilities of learning from cross-lingual demonstration examples using prompt settings. We examine three prompt settings for each language pair: Swahili prompting, where the prompt templates and the example are in the Swahili language; cross-lingual prompting, where examples are in the Swahili language; and English prompting, where the prompt templates and test examples are in the English language. We use the human-translated prompts for Swahili prompting.


\begin{figure*}
\centerline{
\includegraphics[width=\textwidth]{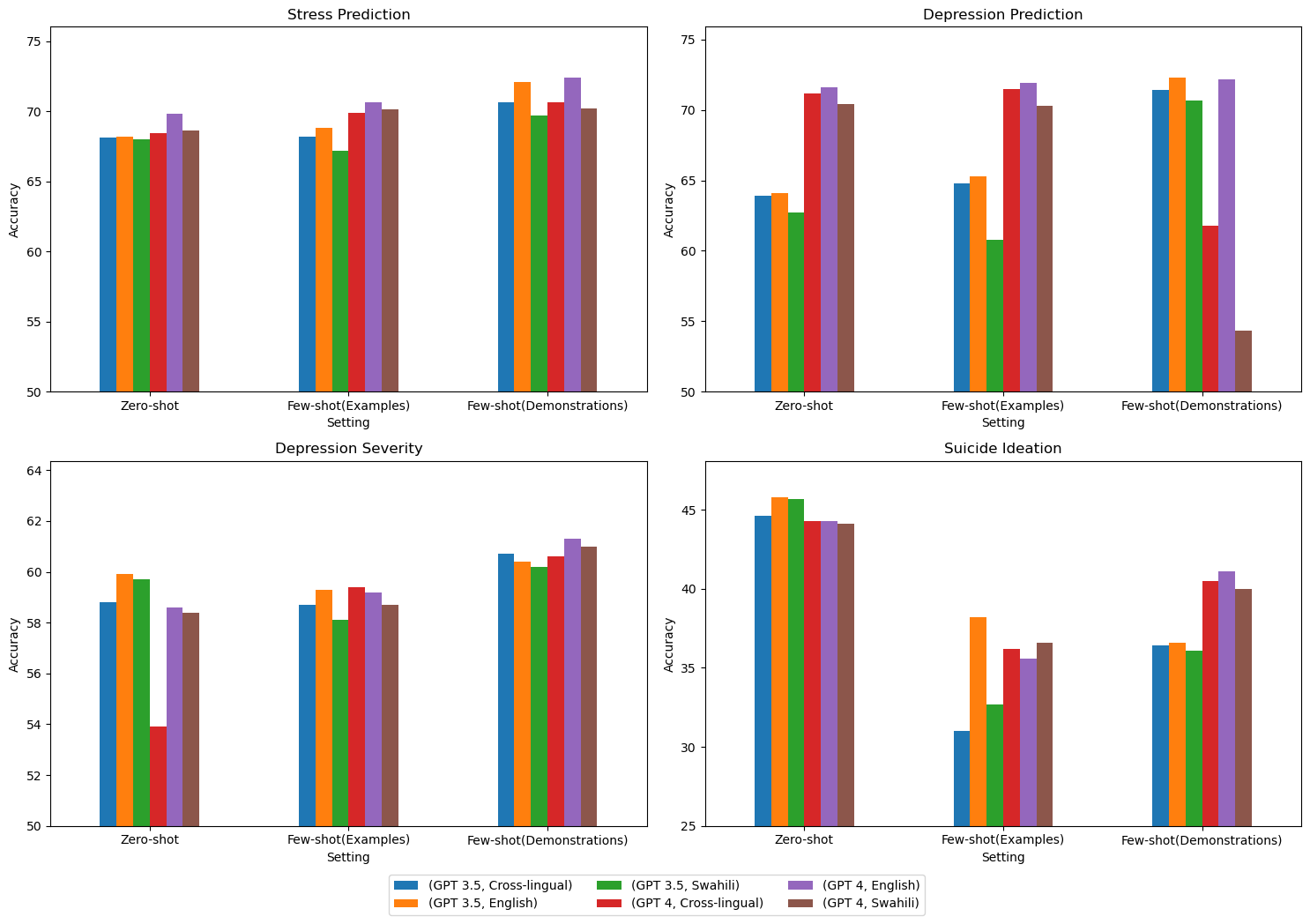} 
}
\caption{Performance accuracy results for the models in three learning scenarios using the three prompt settings: Swahili, Cross-lingual and English, across four mental health tasks.}
\label{fig_learning-scenarios}
\end{figure*}

We observe that the few-shot results with examples are only slightly better across the four tasks than the few-shot results with demonstrations. The results are also sensitive to prompt changes. In zero-shot learning, for example, GPT 3.5 shows improvements fluctuating in handling cross-lingual and Swahili prompts, with a further average of 0.85 increase seen with English prompts, as shown in Fig ~\ref{fig_learning-scenarios}. In contrast, GPT 4 slightly decreases its ability to handle cross-lingual prompts. Still, it demonstrates a remarkable recovery in English prompts, showing improvements of 2.69\% and 1.16\%, respectively. The continuous improvements observed in GPT 3.5 when moving from Swahili to English suggest that context-rich prompts are helpful for in-context learning. GPT 4 shows significant progress, particularly in the depression prediction task (Task 2). However, it exhibits a decline in the stress prediction task (Task 1) when dealing with Cross-lingual prompts, suggesting possible challenges related to language characteristics specific to the task.

Notably, most few-shot (examples only) results are lower than zero-shot results, suggesting that the models cannot use examples effectively under the given conditions. Furthermore, in the few-shot learning, the models tended to replicate the label of the provided example rather than making predictions based on the input data. This further illustrates the difficulty of language models in acquiring knowledge from a few examples in the cross-lingual tasks.

The results also show a decline in performance across all tasks in both models in zero-shot learning. The difference is evident for GPT 4, as it demonstrates a more pronounced decrease in performance for the task of predicting suicidal ideation (Task 4) during in-context learning. This task showed a notably lower performance across all models and prompt strategies than the others. A possible reason for this is that it included longer inputs, which challenged the models' token limits and led to truncation of critical information, resulting in a loss of context and decreased performance.

Both models demonstrate improved performance when using cross-lingual prompts, suggesting that even a few examples help adapt to a cross-lingual environment. More precisely, GPT 3.5 demonstrates clear advantages when used with English prompts. On the other hand, GPT 4, albeit demonstrating improvement in handling prompts from two different languages, encounters a slight setback in tasks such as depression severity (Task 3) and suicide ideation prediction (Task 4).

The English templates demonstrated strong performance across all tasks. Table ~\ref{tab_performance}  presents the average performance accuracy results for the prompt approaches. Swahili's performance was comparatively less effective in the few-shot learning scenario, particularly for GPT 3.5. However, the results show that it performed better than cross-lingual prompts, with an average difference of 0.9\% and -1.9\% over cross-lingual and English prompts.  

As the results show, creating prompt tactics specifically adapted to Swahili's linguistic and cultural aspects could potentially enhance performance in mental health tasks. Nevertheless, we propose that frequently occurring sub-tokens and the quantity of code-switching text in the pre-training data were crucial factors in facilitating cross-lingual prompting. 
\begin{table}[htbp]
  \caption{Average Performance Accuracy Results for the Swahili, Cross-lingual and English Prompt approaches, across all mental health prediction tasks.}
  \label{tab_performance}
  \centering
  \begin{tabular}{|l|l|c|c|c|}
    \hline
    \rowcolor{gray!50}
    \textbf{Setting} & \textbf{Model} & \textbf{Swahili} & \textbf{Cross-lingual} & \textbf{English} \\ \hline
    \multirow{2}{*}{Zero-shot} & GPT 3.5 & 59.0 & 58.9 & 59.5 \\ \cline{2-5} 
     & GPT 4 & 60.4 & 59.5 & 61.1 \\ \hline
    \multirow{2}{*}{\shortstack[l]{Few-shot \\ (Examples)}} & GPT 3.5 & 54.7 & 55.7 & 57.9 \\ \cline{2-5} 
     & GPT 4 & 58.9 & 59.3 & 59.3 \\ \hline
    \multirow{2}{*}{\shortstack[l]{Few-shot \\ (Demonstration)}} & GPT 3.5 & 59.2 & 59.8 & 60.4 \\ \cline{2-5}
     & GPT 4 & 56.4 & 58.4 & 61.8 \\ \hline
  \end{tabular}
\end{table}

In general, high-resource languages with abundant pre-training data and a significant overlap in vocabulary with other languages tend to be more effective universal prompting languages. Additionally, the performance of the Swahili and cross-lingual templates might be partly because the evaluation data for all our mental health tasks were derived from English datasets through translation.

The results indicate that large language models such as GPT 3.5 and GPT 4 can enhance their performance in multilingual scenarios using their in-context learning capabilities. This suggests they possess significant adaptability and potential for transferring language when given a few examples.

\section {Conclusion}
Our study employed a two-fold approach to cross-lingual transfer and adaptation of mental health prediction in a low-resource language. We investigated a model-agnostic meta-learning method and in-context learning with a large language model. We evaluated our framework on three datasets translated into Swahili language across four mental health tasks. Our findings demonstrate that our meta-trained model performs well in zero-shot and few-shot mental health prediction tasks by enhancing cross-lingual transfer and rapid adaptation in the target language, enhancing fine-tuning performance. We showed that cross-lingual meta-training outperforms state-of-the-art fine-tuning baselines in zero-shot and fine-tuning settings.

Furthermore, we used two variants of GPT models and studied their in-context learning capabilities for cross-lingual mental health prediction. We used Swahili, Cross-lingual and English prompt templates. Our analysis showed that the models are capable of cross-lingual transfer as seen with the improvement in Swahili prompts over cross-lingual prompts, leading to a fairly good in-context learning performance. However, it was noted that using these templates in the Swahili language does not always further improve performance, especially when a strong prompting language like English is used. 

Our study suggests that the models can be adapted to understand and respond to cross-lingual environments in the mental health domain. Language-specific prompting strategies can enhance a model's ability to interpret and respond to mental health inquiries accurately. The increase in performance observed with in-context learning highlights the importance of including a few examples and demonstrations in the target language.

One of the major limitations of our study was the limited availability of mental health datasets in the Swahili language, so we used translated datasets. We only investigated the mental tasks in one low-resource language. In the future, it would be pertinent to investigate the impact of multilingual generative models on multiple mental health tasks in various languages to fully capture the linguistic and cultural context inherent in these languages. In addition, given the results observed in our study, we propose a framework that combines the adaptability of meta-learning with the contextual understanding of LLMs for mental health prediction tasks in future.

\bibliographystyle{IEEEtran}
\bibliography{bibliography} 

\vskip 0pt plus -1fil
\begin{IEEEbiography}[{\includegraphics[width=1in,height=1.25in,clip,keepaspectratio]{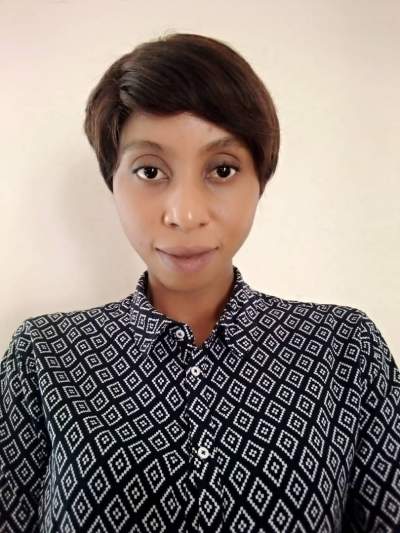}}]{Zita~Lifelo}
received her bachelor’s degree in
Computer Science in 2012 and her Master's degree in Computer Science in 2016 from the Copperbelt University, Kitwe, Zambia.
She is currently a PhD student at University of Science and Technology Beijing, Beijing, China. Her research interests include Social Computing, Natural Language Processing and Affective Computing.
\end{IEEEbiography}

\vskip 0pt plus -1fil
\begin{IEEEbiography}[{\includegraphics[width=1in,height=1.25in,clip,keepaspectratio]{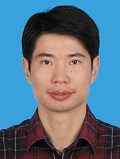}}]{Huansheng Ning}
Received his B.S. degree from Anhui University in 1996 and his Ph.D. degree from Beihang University in 2001. Now, he is a professor and vice dean of the School of Computer and Communication Engineering, University of Science and Technology Beijing, China. His current research focuses on the Internet of Things, General Cyberspace and Metaverse, Smart Education, Cyber-syndrome and Cyber-Health. He has presided many research projects including Natural Science Foundation of China, National High Technology Research and Development Program of China (863 Project). He has published more than 200 journal/conference papers, and authored 5 books. He serves as an associate editor of IEEE Systems Journal (2013-Now), IEEE Internet of Things Journal (2014-2018), and as steering committee member of IEEE Internet of Things Journal (2016-Now).
\end{IEEEbiography}

\vskip 0pt plus -1fil
\begin{IEEEbiography}[{\includegraphics[width=1in,height=1.25in,clip,keepaspectratio]{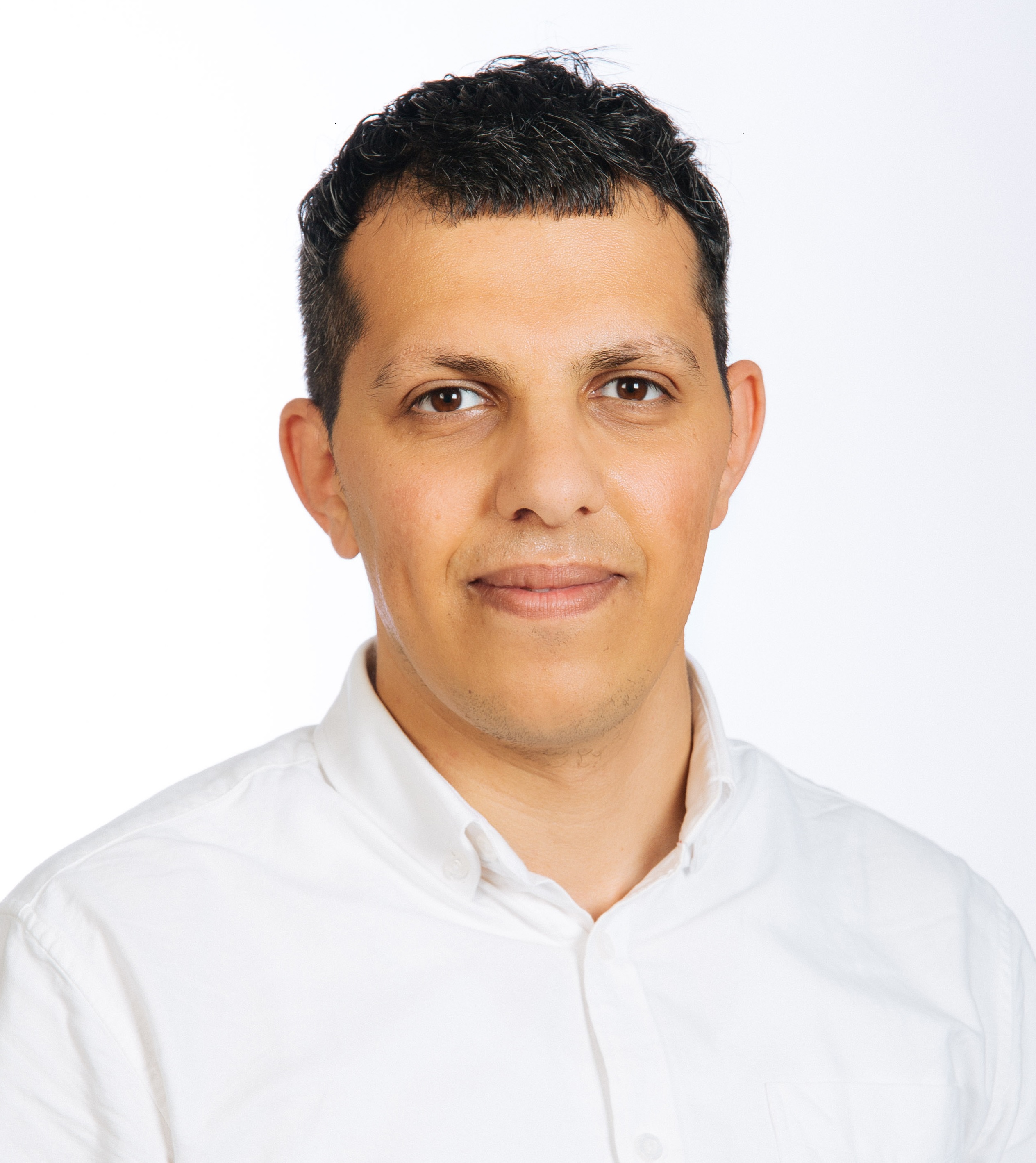}}]{Sahraoui Dhelim} is a senior postdoctoral researcher at University College Dublin, Ireland. He was a visiting researcher at Ulster University, UK (2020-2021). He obtained his PhD degree in Computer Science and Technology from the University of Science and Technology Beijing, China, in 2020. And a Master's degree in Networking and Distributed Systems from the University of Laghouat, Algeria, in 2014. And a B.S. degree in computer science from the University of Djelfa, in 2012. He serves as a guest editor in several reputable journals, including Electronics journal and Applied Science Journal. His research interests include Social Computing, Smart Agriculture, Deep-learning, Recommendation Systems and Intelligent Transportation Systems.
\end{IEEEbiography}

\end{document}